\documentclass[letterpaper]{article} 
\usepackage{aaai23}  
\usepackage{times}  
\usepackage{helvet}  
\usepackage{courier}  
\usepackage[hyphens]{url}  
\usepackage{graphicx} 
\usepackage{natbib}  
\usepackage{caption} 
\usepackage{algorithm}
\usepackage{algorithmic}

\usepackage{newfloat}
\usepackage{listings}
\usepackage{subcaption}
\usepackage{multirow}
\usepackage{booktabs}
\usepackage{bbding}
\usepackage{amsmath}
\usepackage{balance}

\DeclareCaptionStyle{ruled}{labelfont=normalfont,labelsep=colon,strut=off} 
\floatstyle{ruled}
\newfloat{listing}{tb}{lst}{}
\floatname{listing}{Listing}
\title{Accommodating Audio Modality in CLIP for Multimodal Processing}
\author{
    Ludan Ruan,
    Anwen Hu,
    Yuqing Song,
    Liang Zhang,
    Sipeng Zheng,
    Qin Jin\thanks{Corresponding Author.}
}
\affiliations{
    School of Information, Renmin University of China

     \{ruanld,anwenhu,syuqing,zhangliang00,zhengsipeng,qjin\}@ruc.edu.cn
}

\usepackage{bibentry}

\begin{document}

\maketitle

\begin{abstract}
Multimodal processing has attracted much attention lately especially with the success of pre-training. However, the exploration has mainly focused on vision-language pre-training, as introducing more modalities can greatly complicate model design and optimization. In this paper, we extend the state-of-the-art Vision-Language model {CLIP} to accommodate the audio modality for {V}ision-{L}anguage-{A}udio multimodal processing.
Specifically, we apply inter-modal and intra-modal contrastive learning to explore the correlation between audio and other modalities in addition to the inner characteristics of the audio modality.
Moreover, we further design an audio type token to dynamically learn different audio information type for different scenarios, as both verbal and nonverbal heterogeneous information is conveyed in general audios.
Our proposed CLIP4VLA model is validated in different downstream tasks including video retrieval and video captioning, and achieves the state-of-the-art performance on the benchmark datasets of MSR-VTT, VATEX, and Audiocaps. The corresponding code and checkpoints are released at \textit{https://github.com/ludanruan/CLIP4VLA}.
\end{abstract}

\section{Introduction}
Multimodal processing~\cite{Jes_CLIP_MCPR2021,Carion_DERT_ECCV20,Sun_videoBERT_ICCV19} aims to learn the general knowledge across multiple modalities of our daily perception, such as text, vision and audio. Due to the high complexity and high training cost of multimodal alignment, most works focus on the processing of two modalities such as text and vision.
However, only visual and textual information may be insufficient to comprehensively understand a realistic scenario. 
For example, in sports program, the sound of the race start gun and the cheers of the crowd can describe the intensity of the competition even more than the picture, and the narrator's commentary helps the general audience with less sports knowledge to better understand the progress of the game. 
Therefore, it is necessary to equip the video pre-training models with audio modality modeling.

In recent years, pre-training has achieved great success in multimodal processing. For example, Vision-Language~(VL) pre-training models~\cite{Carion_DERT_ECCV20,Sun_videoBERT_ICCV19,Jes_CLIP_MCPR2021} have shown superior performance for understanding tasks such as text-visual retrieval and flexible scalability for generation tasks such as video captioning. Audio pre-training models~\cite{Gong_AST_arxiv21,Alexei_wav2vec2_NIPS20,Chen_WavLM_arxiv21} can represent complex audio information.  
As learning general correlations of vision, text and audio via pre-training from scratch is highly computation costly (e.g. 768 TPU days for VATT~\cite{Akbari_VATT_arxiv21}), one straight-forward idea is to combine the state-of-the-art VL models 
with the pre-trained audio backbones.
However, it faces two main challenges. 
First, the text, vision and audio backbones usually have different model structures, which makes it hard to combine via a unified training strategy.
For example, the audio pre-training models for Automatic Speech Recognition (ASR) normally process audio at the phoneme level, whose parameters are too heavy compared with the VL models.
Second, there is currently no single audio backbone that can fully handle rich and different types of information conveyed in general audios, which can be roughly categorized as verbal information and nonverbal information.
The verbal information refers to the human speech in the video, which delivers linguistic semantics of the video.
The nonverbal information refers to ambient sounds which can reflect natural events occurring in the video, such as raining.
Due to the heterogeneity of these two types of information, the existing audio models usually focus on handling only one type.
However, both types of information are indispensable for the comprehensive video understanding. It is naive and cumbersome to apply multiple audio backbones to encode the two types of audio information respectively.

\begin{figure*}[t]
  \centering
\includegraphics[width=0.85\textwidth]{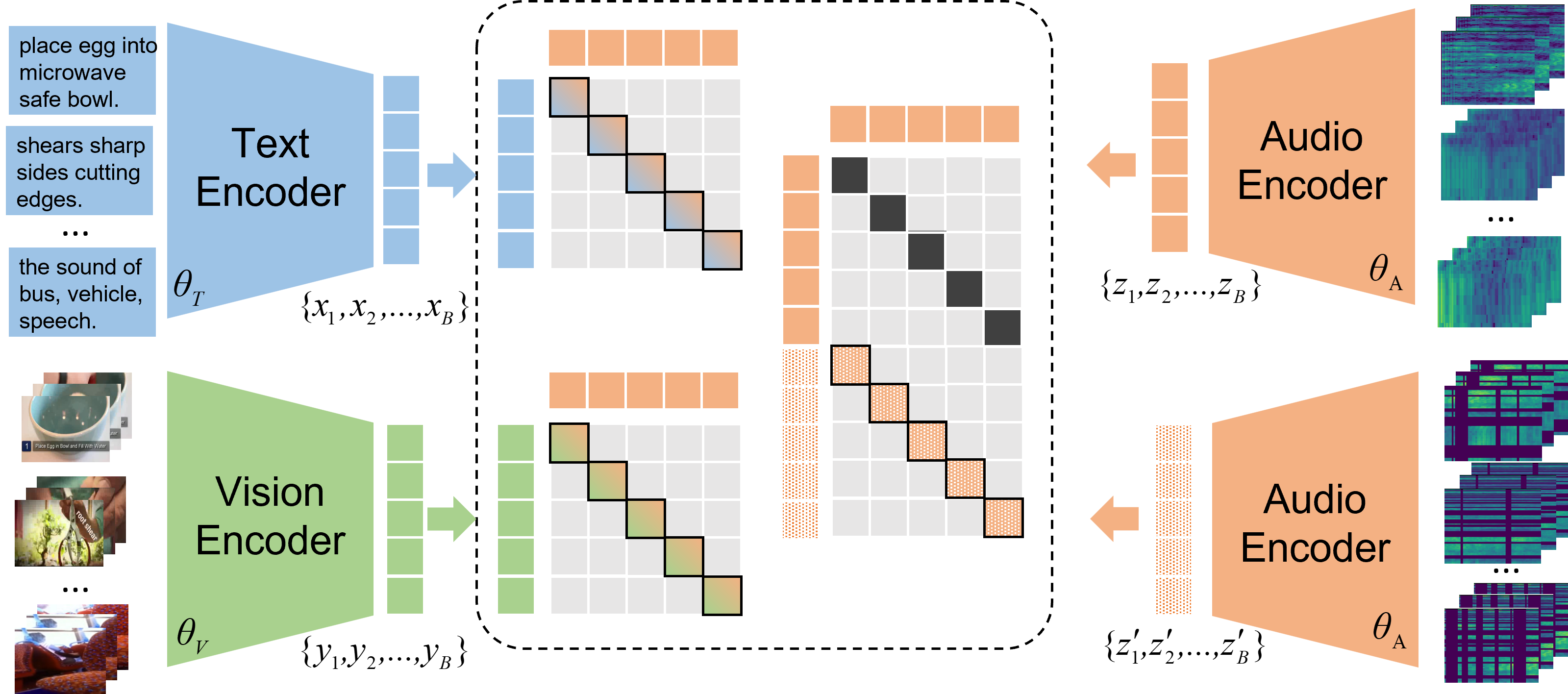}
  \caption{An overview of our CLIP4VLA model, which consists of three backbones: Text Encoder, Vision Encoder, and Audio Encoder. After encoding a batch of text features $\{x_1,x_2,\cdots,x_B\}$, vision features $\{y_1,y_2,\cdots,y_B\}$ and audio features $\{z_1,z_2,\cdots,z_B\}$, we pre-train the model with four kinds of contrastive learning objectives for text-audio, video-audio, augmented\_audio-original\_audio respectively. The black squares are not included in the calculation}
  \label{fig:model_all}
\end{figure*}

To tackle the above two challenges, in this paper, we propose \textbf{CLIP4VLA}~(\textbf{CLIP} for \textbf{V}ision, \textbf{L}anguage and \textbf{A}udio), which extends CLIP to accommodate the audio modality with unified tri-encoder structure for multimodal processing.
Specifically, we employ the state-of-the-art VL model CLIP~\cite{Jes_CLIP_MCPR2021} as the vision and text encoders, and propose an audio encoder with the same architecture as the vision encoder to ensure the training consistency and efficiency.
To simultaneously encode both verbal information and nonverbal information from the audio track of videos, we design an audio type token to dynamically control the learned information type.
During pre-training, we apply both inter-modal and intra-modal contrastive learning to learn the correlation across audio modality with other modalities and the inner characteristics of the audio modality.
To better utilize the multimodal representations learned by CLIP4VLA, we further explore different modality fusion methods for video-text downstream tasks on various datasets. CLIP4VLA is demonstrated to be effective on both retrieval and captioning tasks, requiring much less hardware resource and training time.

Our contributions can be summarized as follows:
\parskip=0.1em
\begin{itemize}%
\item We propose CLIP4VLA for learning correlation across textual, visual and audio information in videos by accommodating the audio encoder in CLIP.
\item To fully exploit the rich audio information in videos, we propose to explicitly encode both verbal information and nonverbal information with audio type tokens.
\item We design intra-modal and inter-modal contrastive learning for  pre-training CLIP4VLA and explore multiple modality fusion methods for video downstream tasks.
\item Our model achieves the state-of-the-art performance in retrieval and captioning tasks on the benchmark datasets of MSR-VTT, VATEX, and Audiocaps.
\end{itemize}

\section{Related Work}
\subsection{Audio Pre-training}
Audio pre-training works aim to well represent nonverbal information in ambient sound \cite{Gemmeke_Audioset_ICASSP17,Gong_AST_arxiv21,Guzhov_AudioCLIP_arxiv21,Andrey_ESResnet_ICPR20,Wu_Wav2CLIP_arxiv21} or verbal information in human speech \cite{Liu_TERA_ACM21,Tang_DeCEMBERT_NAACL21,Chung_APC_Interspeech19,Alexei_wav2vec2_NIPS20,Hsu_HuBERT_ACM21,Chen_WavLM_arxiv21} .  
For nonverbal information encoding, recent works~\cite{Guzhov_AudioCLIP_arxiv21,Andrey_ESResnet_ICPR20,Wu_Wav2CLIP_arxiv21,Gong_AST_arxiv21} prove that audio representation learning can benefit from other modalities (i.e. images) by transfer learning.
To encode verbal information, self-supervised methods are always utilized to learn inherent characteristic, ranging from auto-regressive learning  \cite{Chung_APC_Interspeech19,Liu_NPC_arxiv20,Liu_TERA_ACM21} to contrastive learning~\cite{Alexei_wav2vec2_NIPS20,Oord_CPC_arxiv18,Ling_Decoar_ICASSP20}. 
Furthermore, wav2vec2.0~\cite{Alexei_wav2vec2_NIPS20}, HuBERT~\cite{Hsu_HuBERT_ACM21}, WavLM~\cite{Chen_WavLM_arxiv21} demonstrate that self-supervised learning with a large amount of unlabeled data could boost the model's performance on semantic related tasks~(i.e. ASR) and decrease the demand of labeled data.
Our CLIP4VLA has two changes compared with the previous audio pre-training works: Firstly, previous works mainly focus on audio encoding and ignore cross-modality understanding, while CLIP4VLA enhances audio representation by both self-supervised learning and cross-modal alignment. Secondly, previous works only focus on one specific type of audios while CLIP4VLA extract both verbal and nonverbal information for general video understanding.

\begin{figure*}[t]
  \centering
  \includegraphics[width=0.95\textwidth]{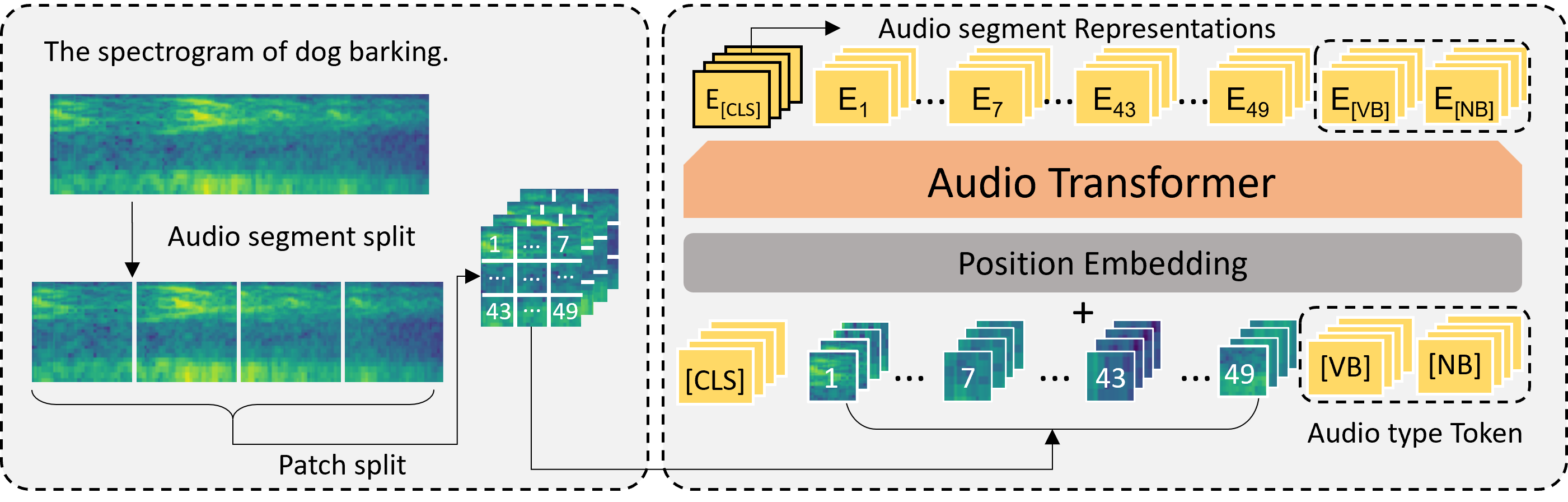}
  \vspace{-5pt}
  \caption{ An overview of the audio encoding.  The audio spectrogram is first split into segments along temporal dimension without overlap, the spectrogram of each audio segment is then split into a patch sequence of 7 $\times$ 7 
  without overlap. After flattening, A [CLS] token and an audio type embedding are added to the start and the end of the patch sequence respectively. Each patch embedding is added with a learnable positional embedding and then fed into the Audio encoder, which keeps the same structure as the visual encoder. The output of [CLS] is used as the final audio segment representation }
  \label{fig:model_audio}
\end{figure*}

\subsection{Video-Text Pre-training}

Most video-text pre-training works~\cite{Sun_videoBERT_ICCV19,Tang_DeCEMBERT_NAACL21,Xu_VLM_ACL21,Lei_ClipBert_CVPR21,Sun_CBT_ECCV20,Zhu_ActBERT_CVPR20,luo_univl_arxiv20} focus on the vision-text alignment in videos. VideoBERT~\cite{Sun_videoBERT_ICCV19} and CBT~\cite{Sun_CBT_ECCV20} are pioneering works to explore Video-Language representation by self-supervised learning.
For fine-grained multimodal understanding, HERO~\cite{Li_HERO_EMNLP20} designs a temporal-specific proxy task and  UniVL~\cite{luo_univl_arxiv20} designs a generation proxy task.
ClipBERT~\cite{Lei_ClipBert_CVPR21} further explores an end-to-end manner by inputting sparse sampled frames from video clips rather than extracted video features from pre-trained backbones~\cite{Xie_S3D_ECCV18}. These works well explore the correlation between vision and text modalities but ignore audio information in videos. 

Recently, some works \cite{Alayrac_MMV_NIPS20,Akbari_VATT_arxiv21,Liu_OPT_arxiv21} try to incorporate audio modality during pretraining for tri-modal understanding. OPT~\cite{Liu_OPT_arxiv21} focuses on the speech of image descriptions, which is greatly different from the audio in general videos.  
To encode general videos, VATT~\cite{Akbari_VATT_arxiv21} explores representing all three modalities with one modality-agnostic encoder. 
Our work also focuses on general videos for broader application and there are major two differences.
Firstly, VATT is trained from scratch with heavy computation load while our model learns triple-modal correlation based on existing VL pe-trained model.
Secondly, VATT does not distinguish verbal and nonverbal information in audios while CLIP4VLA respectively learns their correlation with other two modalities from different types of videos.

\section{Method}
In this section, we describe the proposed CLIP4VLA model and the multimodal contrastive learning objectives for pre-training in details. 
Given a batch of videos 
and their corresponding descriptions, we first extract audios from the videos and formulate the video batch, audio batch and text batch as $V$, $A$ and $T$ respectively.
The target of our CLIP4VLA model is to learn rich semantic representations for the three modalities, so that the corresponding video, audio and text with similar semantics can be embedded close to each other though in different modalities, while those with different semantics be embedded further away.

With the multimodal representations fully learned, we adapt the model on different downstream tasks including cross-modal retrieval and multimodal captioning to verify the effectiveness of our CLIP4VLA model.

\subsection{Model Structure}
As illustrated in Figure~\ref{fig:model_all}, our proposed CLIP4VLA model consists of three backbones to handle the textual, visual and audio signals respectively.
The details of the audio processing and audio backbone structure are illustrated in Figure~\ref{fig:model_audio}.

\begin{figure*}[t]
  \centering
  \includegraphics[width=0.9\textwidth]{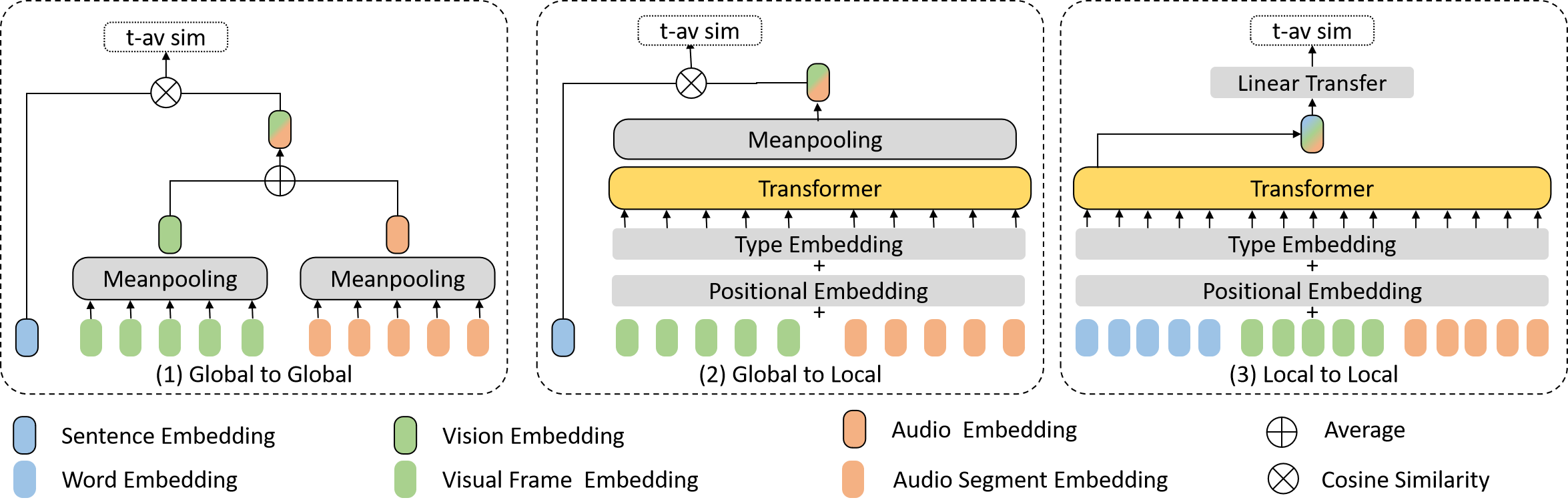}
  \caption{An overview of different modal fusion methods. We explore 3 methods for video retrieval, which includes (1) Global to Global, (2) Global to Local, (3) Local to Local }
  \label{fig:model_fusion}
\end{figure*}

\noindent\textbf{Text \& Vision Encoder.}
We employ CLIP~\cite{Jes_CLIP_MCPR2021} as our text and vision encoders to encode text input $T$ and vision input $V$.
Each $t_i \in T$ is first tokenized into a token sequence and then added with a start token [SOS] and an end token [EOS], denoted as $\{t_i^1 , t_i^2,\cdots, t_i^{L_T} \}$.  
After text encoding, outputs of each token are collected as word-level representations $\{x_i^1,x_i^2,\cdots,x_i^{L_T}\}$. Following CLIP, we choose the output of [EOS] token as the global text representation of $t_i$, denoted as $x_i^g$.

For visual information in videos, we uniformly sample $L_V$ frames from $v_i \in V$ in the temporal dimension as the vision sequence $\{v_i^1,v_i^2,\cdots, v_i^{L_V}\}$.
Specifically, each frame is split into a sequence of patches without overlap and then added with a [CLS] token.
During vision encoding, patch sequence of each frame is independently fed into the vision encoder to model the spatial relationship between patches. The final output of the [CLS] token is chosen as the vision representation of each video frame.
Finally, for the vision sequence $\{v_i^1,v_i^2,\cdots, v_i^{L_V}\}$, we acquire frame-level vision representations $y_i=\{y_i^1,y_i^2,\cdots, y_i^{L_V}\}$. By average pooling of $y_i$, we get a global vision embedding, denoted as $y_i^g$.

\noindent\textbf{Audio Encoder\label{sec:audio}}
Well-trained specialists could infer ambient events or human voice by watching spectrograms. Thus it is also possible for machine to encode audio information with visual spectrograms as inputs.
To keep architecture consistency across different modalities, we design our audio encoder with the same model structure as the vision encoder. 
To process audios similarly as the visual signals, the first thing to do is to transfer the 1-dimensional long audio $a_i \in A$ into the image format, a matrix in the shape of 224$\times$224$\times$3 in this paper.
To be specific, we convert the audio waveform into 224-dimensional log Mel filterbank~(fbank) features with 32ms Hamming window every 8ms.
In this way, a t-second audio stream will be transferred into a spectrogram in the shape of 125t $\times$ 224. 
We cut the spectrogram into k$\times$224$\times$224 along the temporal dimension without overlap, and pad with zeros if the last part is less than 224.
Therefore, a t-second audio will be finally transferred into $\lceil \frac{t}{1.792} \rceil$ frames $\{ a_i^1,a_i^2,\cdots, a_i^{L_A}\}$ with $a_i^j \in R^{224 \times 224 \times 3}$.
Then the normalized segment frames can be encoded similarly with frame images.
The final sequence of audio segment representations is denoted as $z_i=\{z_i^1,z_i^2,\cdots,z_i^{L_A}\}$.  We also get a global audio embedding  by average pooling of $z_i$, denoted as $z_i^g$.

\noindent\textbf{Audio Type Token}
As we consider roughly two types of information in the audios of common videos~(\textbf{V}er\textbf{B}al information and \textbf{N}onver\textbf{B}al information), we design audio type tokens to effectively control which type of features the audio encoder tends to generate.
To be specific, after flattening patches of each audio segment, an audio type token [VB]/[NB] is added at the end of the patch sequence according to different application scenarios. 
For example, for dialogue or commentary, the audio type token [VB] could be used to encode the verbal information.
While for natural activities/events, where the nonverbal information is more important, the [NB] token can be used as the control signal to extract the audio features from the nonverbal aspect.
Furthermore, for the complex scenarios where both verbal information and nonverbal information are crucial, these two types of embeddings can also be combined for better video understanding.
During pre-training, we set the audio type token according to the characteristics of audio pre-training datasets.
During fine-tuning or testing, we add both audio type tokens at the end of the flatten patch sequences to flexibly extract both verbal and nonverbal information.

\subsection{Pre-training \label{sec:pre-training_objectives}}
In this section, we introduce pre-training objectives of our CLIP4VLA model.
To learn semantic representations of text, vision and audio, we explore contrastive learning from two perspectives: inter-modal and intra-modal.
The inter-modal contrastive learning is designed to learn the correlation between audio modality and text/vision modality.
The intra-modal contrastive learning aims to learn the inherent characteristics of the audio modality.
We choose the NCE loss~\cite{Jozefowicz_NCEloss_arxiv16} for both inter-modal and intra-modal contrastive learning.

\noindent\textbf{Inter-modal Learning}
During inter-modal learning, which learns cross-modal alignments between text, vision and audio, positive pairs of cross-modal representations should be closer than negative ones. In this work, we construct negative pairs of cross-modal representations within a mini-batch.
With global embeddings of text, vision and audio modalities, we compute the cosine similarity matrix in $B \times B$ for text-audio pairs and vision-audio pairs within a mini-batch, where $B$ is the batch size.
Since the vision and text encoders have been well pre-trained to learn vision-language alignment, we mainly train the audio encoder by maximizing the cosine similarity of $B$ positive pairs while minimizing the cosine similarity of the $B^2 - B$ negative pairs. The symmetric cross entropy loss 
is calculated as follows:
\begin{align}
     &{\rm NCE}_\textit{at} = \frac{1}{B}\sum_{i}^{B}{\log\frac{\exp(z_i^g \cdot x_i^g)}{\sum_{j}^{B}{\exp(z_i^g \cdot x_j^g)}}} \\
     &{\rm NCE}_\textit{av} = \frac{1}{B}\sum_{i}^{B}{\log\frac{\exp(z_i^g \cdot y_i^g)}{\sum_{j}^{B}{\exp(z_i^g \cdot y_j^g)}}}
\end{align}
where $x^g$, $y^g$, $z^g$ refer to global embbeddings of text, vision and audio modalities.

\begin{table*}[htbp]
\centering
  \begin{tabular}{lcccccccccc}
    \toprule
    ~ & \multicolumn{4}{c}{MSR-VTT} && \multicolumn{4}{c}{VATEX}\\
    \cline{2-5} \cline{7-10} 
    Model& R@1 & R@5 & R@10 & MedianR & & R@1 & R@5 & R@10 & MedianR \\
     \midrule
      W2VV++~\cite{Li_W2VV_MM19} &18.9 &45.3& 57.5  & - &&  34.3& 73.6& 83.7& -\\
     CE~\cite{Liu_CE_BMVC19}& 20.9&48.8 &62.4  & 5.0 &&47.9& 84.2& 91.3& 2.0\\
     MMT~\cite{Gabeur_MMT_ECCV20}&26.6& 57.1& 69.6&-&&-&-&-&-\\
     HGR~\cite{Chen_HGR_CVPR20} &- &- & - & - && 35.1& 73.5& 83.5& 2.0\\
     SSB~\cite{Patrick_SSB_ICLR_21} & 30.1& 58.5& 69.3& 3.0 && 45.9& 82.4& 90.4& 1.0\\
     \hline
      UniVL~\cite{luo_univl_arxiv20} &20.6& 49.1& 62.9& 6.0&&-&-&-&-\\
     ClipBERT~\cite{Lei_ClipBert_CVPR21}&22.0&46.8&59.9&6.0&&-&-&-&-\\
     VLM~\cite{Xu_VLM_ACL21}&28.1&55.5&57.4&4.0&&-&-&-&-\\
     \hline
    CLIP & 31.2 & 53.7 & 64.2 & 4.0 &&39.7 &72.3& 82.2&- \\
    CLIP-FRL~\cite{Chen_CLIPFRL_ICCV21} & 38.2&66.0& 75.7  & - && 47.1& 82.3& 90.6 & -\\
     CLIP4Clip~\cite{Luo_CLIP4Clip_arxiv21}&44.5&71.4&81.6&2.0  && 55.9 & 89.2 & 95.0 &1.0 \\
     CLIP2Video~\cite{Fang_CLIP2Video_arxiv21}&45.6& 72.5& 81.7& 2.0 &&57.3& 90.0 &95.5& 1.0 \\
     \midrule
      CLIP4VLA & \textbf{46.2} & \textbf{73.5} & \textbf{83.5} & 
      \textbf{2.0} &&\textbf{63.5}&\textbf{91.5}&\textbf{95.9}&\textbf{1.0} \\ 
    \bottomrule
\end{tabular}
\vspace{-5pt}
\caption{Video Retrieval Performance on MSR-VTT-1kA and VATEX}
\label{tab:performance_retrieval}
\end{table*}

\begin{table}[t]
\centering
\scalebox{0.9}{
  \begin{tabular}{lccccc}
    \toprule
     Model& Modality  &R@1& R@5 & R@10& MedianR \\
     \midrule
    VGGish &A& 18.5 & - & 62.0 &-\\
    VGGSound &A&22.4 & - & 69.2 &-\\
    MoEE &A&22.5&-&69.5& -\\
    CE &A& 23.1&56.2 & 70.7 &4.0\\
    CLIP4VLA& A&\textbf{28.4}&\textbf{60.9}&\textbf{76.2}&4.0\\
    \midrule
    CE&AV& 28.0&-& 80.4&-\\
    
      CLIP4VLA &AV& \textbf{33.6}& \textbf{68.1} & \textbf{82.3}&3.0 \\
    \bottomrule
\end{tabular}}
\vspace{-5pt}
\caption{Retrieval Performance Comparison on Audiocaps} 

\label{tab:performance_retrieval_audiocaps}
\end{table}

\noindent\textbf{Intra-modal Learning}
To enhance the information representation ability of audio encoder, we further optimize it with intra-modal self-supervised learning.
We first augment the audio $a_i$ to $\hat{a_i}$ by randomly masking the audio spectrograms along both channel and temporal dimension~\cite{Liu_TERA_ACM21}. 
To be specific, we randomly sample the start step along the channel step and the time step with probability of 5\% and 15\% respectively, then we mask the subsequent 10 consecutive steps from the start step. Overlap is allowed in the masking. 
The original audio $a_i$ and its augmented version $\hat{a_i}$ then can be seen as a positive pair for the contrastive learning.
Similar to the inter-modal NCE loss, other masked audios within a mini-batch are negative samples for $a_i$. The symmetric cross entropy loss is calculated as follows:
\begin{align}
     {\rm NCE}_{\textit{a}\hat{a}} & =  \frac{1}{B}\sum_{i}^{B}{\log\frac{\exp(z_i^g \cdot \hat{z_i}^g)}{S}} \\
     S & = \sum_{j}^{B}{\exp(z_i^g \cdot \hat{z_j}^g)} + \sum_{k\neq i}^{B}{\exp(z_i^g \cdot z_k}^g)
\end{align}
where $\hat{z_i}^g$ is the global embedding of masked audio $\hat{a_i}$.

The final pre-training loss for CLIP4VLA is the sum of  inter-modal NCE and intra-modal NCE objectives:
\begin{align}
\mathcal{L}={\rm NCE}_\textit{at}+{\rm NCE}_\textit{av}+{\rm NCE}_{\textit{a}\hat{a}}
\end{align}

\subsection{Fine-tuning}
To verify the effectiveness of the learned representations for text, vision and audio, we fine-tune the CLIP4VLA model for multiple downstream tasks.

\noindent\textbf{Fine-tuning for Video Retrieval}
Video Retrieval aims to search the target video based on a video caption as the retrieval query.
Without encoding audio information, existing video retrieval works \cite{Liu_CE_BMVC19,Miech_moee_arxiv18,Chen_HGR_CVPR20} only focus on the matching between text and vision modality.  
Benefiting from the tri-modality encoding ability of CLIP4VLA, we fully explore both vision and audio information in the video for text-video retrieval.
Since there are three modalities involved in this task, effective multimodal fusion is important.
In this paper, we explore three multimodal fusion approaches for text-video retrieval, including (1) \emph{Global to Global}, (2) \emph{Global to Local}, and (3) \emph{Local to Local}. As illustrated in Figure~\ref{fig:model_fusion},
the \emph{Global to Global} approach directly calculates similarity based on the global embeddings of vision and audio modalities via mean pooling.
For the~\emph{Global to Local} approach, we apply a Video Temporal Encoding Module~(N-layer transformer) to encode temporal relevance of vision and audio modalities, and calculate the similarity between text feature and fused video feature. For the~\emph{Local to Local} approach, we apply a Fine-grained Cross-modality Fusion Module (N-layer transformer) to further exploit the fine-grained correlation of text to vision and audio modalities. We analyze these multimodal fusion methods  in the supplementary material.

\noindent\textbf{Fine-tuning for Video Captioning}
Besides the video retrieval task, Video Captioning~\cite{Zhang_ORG_CVPR20,Lin_APML_AAAI21,Wang_CMG_arxiv21} is another challenging task on video understanding, which aims to generate fluent natural language description of video contents. 
To conduct sentence generation, we introduce a Multimodal Caption Generator (N-layer transformer encoder) upon CLIP4VLA.
At the $t^{th}$ decoding step, we feed previous generated words, vision frames and audio segments into CLIP4VLA. After intra-model encoding with three encoders, we concatenate the fine-grained features to construct multimodal sequence $U_i$.
 Input the sequence into  Multimodal Caption Generator, the $t^{th}$ word is predicted as follows:
\begin{align}
    H_i &= {\rm MCG}(U_i), \\
    p_i^{t} &= {\rm softmax}(f(h_i^t)), \quad h_i^t \in H_i,
\end{align}
where MCG refers to the  Multimodal Caption Generator, $f(\cdot)$ is the linear output layer, $p_i^{t}$ is the predicted probabilities over the whole vocabulary size.

\begin{figure*}[t]
  \centering
\includegraphics[width=0.85\textwidth]{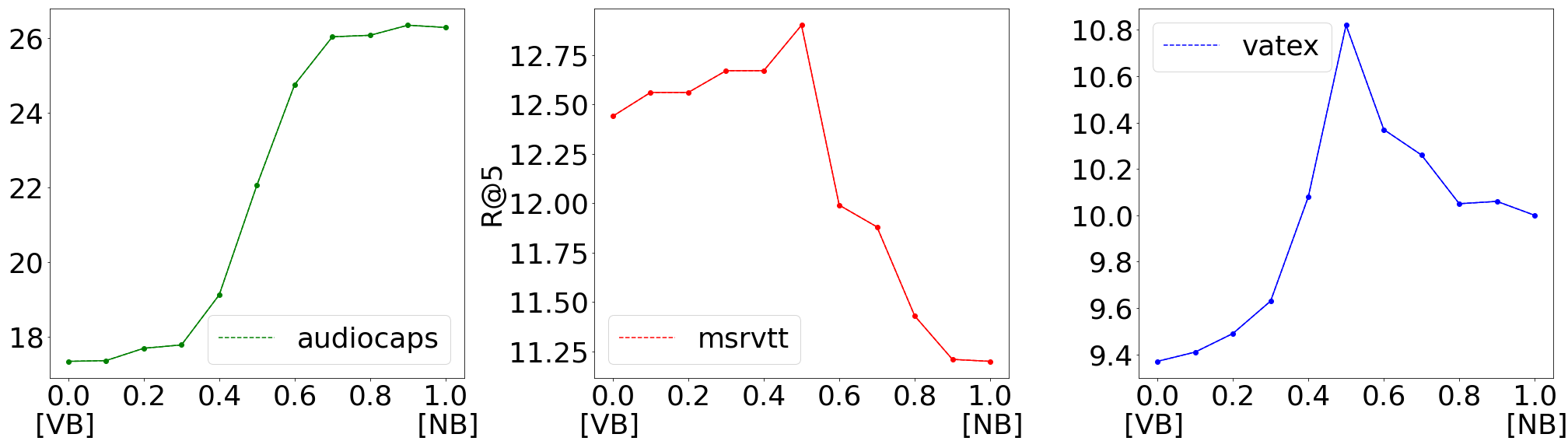}
  \caption{ The zero-shot text-audio retrieval performance on different datasets with different audio type token. The x-axis represents the mixing ratio of [VB] and [NB] type embeddings, and the y-axis represents the retrieval performance}
  \label{fig:control_token}
\end{figure*}

\begin{table}[t]
\centering
\scalebox{0.95}{
  \begin{tabular}{lccccc}
    \toprule
     Model  & BLUE4& METEOR & ROUGE& CIDER \\
     \midrule
     ORG-TRAL&43.6&28.8&62.1&50.9\\
     SemSynAN &44.3&  28.8&62.5& 50.1\\
     APML &41.9&29.9&62.6&49.8\\
     UniVL&42.2& 28.8&61.2 & 49.9 \\
     CMG &43.7&29.4&62.8& 55.9\\
     Clip4Caption &46.1&30.7&\textbf{64.8}&57.7\\
      CLIP4VLA & \textbf{46.7} &\textbf{31.1}& 64.4&\textbf{58.0}
      \\ 
    \bottomrule
\end{tabular}
}
\vspace{-5pt}
\caption{Captioning Performance on MSR-VTT}
\label{tab:performance_caption_msrvtt}
\end{table}

\begin{table}[t]
\centering
  \begin{tabular}{lccccc}
    \toprule
     Model  & BLUE4& METEOR & ROUGE& CIDER  \\
     \midrule
     Shared E &28.4& 21.7& 47.0 & 45.1\\
     Shared E-D &27.9& 21.6&  46.8 & 44.2 \\
     ORG-TRAL&32.1&22.2&48.9& 49.7\\
     SCST-C-B-F&33.3&22.8&49.6&54.6\\
      CLIP4VLA &\textbf{36.4}&\textbf{25.0}&\textbf{54.7}&\textbf{59.7}\\
    \bottomrule
\end{tabular}
\vspace{-5pt}
\caption{Captioning Performance on VATEX }
\label{tab:performance_caption_vatex}
\end{table}

\section{Experiments}
\subsection{Experiment Settings}
 We first pre-train our proposed CLIP4VLA on large scale datasets including Howto100M~\cite{Miech_Howto100m_ICCV19} and Audioset~\cite{Gemmeke_Audioset_ICASSP17}, then fine-tune it for the retrieval and  captioning tasks on three datasets: MSR-VTT~\cite{Xu_MSRVTT_CVPR16}, VATEX~\cite{Wang_Vatex_ICCV19}, Audiocaps~\cite{Kim_audiocaps_NAACL19}.
The evaluation metrics are Recall@n~(R@n) and Median R for retrieval tasks, 
and BLUE-n, METEOR, ROUGE, CIDER for captioning tasks. 

\noindent\textbf{Pre-training Datasets} 
Our pre-training data includes instructional video dataset Howto100M~\cite{Miech_Howto100m_ICCV19} and event video dataset Audioset~\cite{Gemmeke_Audioset_ICASSP17}.
To enable our audio encoder to distinguish verbal and nonverbal audio information, we choose [VB] as the audio type token for Howto100M and [NB] for Audioset, respectively.
More details of data processing can be found in the supplementary material.

\noindent\textbf{Fine-tuning Datasets}
We evaluate the pre-trained CLIP4VLA on retrieval and captioning benchmarks, including MSR-VTT~\cite{Xu_MSRVTT_CVPR16}, VATEX~\cite{Wang_Vatex_ICCV19} and Audiocaps~\cite{Kim_audiocaps_NAACL19}. After filtering out the silent videos, 
MSR-VTT  remains 7867 and 884 videos for training and testing on the retrieval task, and 5867, 448, and 2617 videos for training, validation, and testing on the captioning task. 
VATEX  remains 24667, 1427, and 1421 videos for training, validation, and testing on retrieval, and 24667, 2845, and 5698 videos for training, validation, and testing on captioning.
Audiocaps keeps 49712, 495, and 967 videos for training, validation, and testing on the retrieval task.
For training cost comparison with previous work, we further measure our model on event classification datasets of UCF101~\cite{Khurram_UCF101_2012} and ESC50~\cite{Piczak_ESC50_mm15}. The former one contains 13K videos of 101 action classes, and the latter one contains 2K audio clips of 50 classes.

\begin{table*}[t]
\centering
  \begin{tabular}{c|l|cccccccccc}
    \toprule
    ~& &\multicolumn{4}{c}{MSR-VTT} && \multicolumn{4}{c}{VATEX}  \\
    \cline{3-6} \cline{8-11} 
    & ~~~~ Components & R@1 & R@5 & R@10 & MedianR  & & R@1 & R@5 & R@10 & MedianR  \\
     \midrule
     1 & Scratch &2.5&7.8&11.1&  124.0&&1.9&6.0&9.4&156.0\\
     2 & +Initial&3.6& 11.6& 16.5& 136.5&&3.7&12.5&19.3&  72.0\\
     3 & ~~+Inter-modal NCE&6.8&  15.4&21.6&81.5&&7.0&19.7&27.5&  40.0\\
     4 & ~~~~+Intra-modal NCE& 10.5&25.4&37.5&21.0&&9.3&24.9&34.2&\textbf{26.0}\\
     5 & ~~~~~~+Audio Type Token&\textbf{10.6}&  \textbf{26.5} &\textbf{38.0}&\textbf{19.0}&&\textbf{9.9}&\textbf{25.4}&\textbf{34.3}&29.0\\
    \bottomrule
\end{tabular}
\vspace{-5pt}
\caption{Impact of key components for audio retrieval on MSR-VTT \& VATEX}
\label{tab:abalation_components_retrieval2}
\end{table*}

\begin{table*}[htbp]
    \centering
    \begin{tabular}{l|ccccccccccccc}
    \toprule
    Model&&Training Cost&&Batch Size&& Training Param&& ESC50(A)&& UCF101(V\& A)\\
    \toprule
       VATT-Medium  && 512$\sim$768 TPU days&& 2048 && 264M && 84.7&& 89.6\\
       CLIP4VLA && 48 V100 days && 256 && 88M && 86.8 && 91.9\\
        \bottomrule
    \end{tabular}
    \vspace{-5pt}
    \caption{The Comparison of Training Cost and event classification performance of VATT and CLIP4VLA on ESC50 and UCF101~(audio features of ESC50, vision and audio features of UCF101)}    
    \label{tab:training_cost}
\end{table*}

\subsection{Comparison with the State-of-the-arts}
\noindent\textbf{Video Retrieval}
To demonstrate the effectiveness of our proposed CLIP4VLA, we first evaluate it for video retrieval on three benchmarks. Baselines could be grouped into three categories, corresponding to the three blocks in the table: 1) \textit{Classical Retrieval Methods}: W2VV++~\cite{Li_W2VV_MM19}, CE~\cite{Liu_CE_BMVC19}, HGR~\cite{Chen_HGR_CVPR20}, MMT~\cite{Gabeur_MMT_ECCV20}, SSB~\cite{Patrick_SSB_ICLR_21}, MoEE~\cite{Miech_moee_arxiv18}; 2) \textit{Pre-training based Methods}: UniVL~\cite{luo_univl_arxiv20}, ClipBERT~\cite{Lei_ClipBert_CVPR21}, VLM~\cite{Xu_VLM_ACL21}; 3) \textit{CLIP-based Methods}: CLIP~\cite{Jes_CLIP_MCPR2021}, CLIP-FRL~\cite{Chen_CLIPFRL_ICCV21}, CLIP4Clip~\cite{Luo_CLIP4Clip_arxiv21}, CLIP2Video~\cite{Fang_CLIP2Video_arxiv21}. 
Our model understands videos according to both vision and audio information. However, on visual-centric video datasets MSR-VTT and VATEX, not all videos contain audio information. To deal with the missing modality problem for each silent video during testing,  we pair it with the audio of the most similar video, which is chosen from corresponding training set according to cosine similarity of global vision features. 
As shown in Table~\ref{tab:performance_retrieval}, firstly, our model CLIP4VLA achieves state-of-the-art performance on both MSR-VTT and VATEX datasets. 
Secondly, CLIP-based methods significantly outperform other baselines, which shows video understanding can benefit a lot from large-scale image-text pre-training.
Thirdly, with audio content as extra input, our CLIP4VLA achieves better performance than other CLIP-based methods. This indicates that our model could well encode the correlation across text, vision and audio modality.

Besides visual-centric datasets, we also evaluate our model on audio-centric video dataset Audiocaps. As shown in Table~\ref{tab:performance_retrieval_audiocaps},
either with only audio representations or both audio and vision representations of videos, our model achieves state-of-the-art video retrieval performance on Audiocaps. 
What's more, CLIP4VLA  with both audio and vision information outperforms the one with only audio information. This indicates that our model could better understand audio-centric videos by leveraging vision information.  

\noindent\textbf{Video Captioning}
We further validate the adaptability of CLIP4VLA to video captioning task on MSRVTT and VATEXT. As shown in 
Table~\ref{tab:performance_caption_msrvtt} and Table~\ref{tab:performance_caption_vatex}, our model achieves state-of-the-art captioning performance on both datasets as well. This indicates that our model also possesses good caption generation capability by leveraging well-aligned multimodal representations.

\subsection{Ablation Study}
\noindent\textbf{Audio Type Token \label{sec:control_token}}
To verify the validity of our proposed audio type token for different kinds of audio information encoding, we conduct the experiment to compare the audio retrieval performance when adjusting the mixing ratio of the two type embeddings of [NB] and [VB].
As shown in Figure~\ref{fig:control_token}, with the mixing ratio of [NB] embedding increased, the audio retrieval result on the Audiocaps dataset is significantly improved, because most of the audios in Audiocaps dataset are ambient sound.
However, for the video datasets MSR-VTT and VATEX, the best results are yielded when the [NB] and [VB] embeddings are mixed with a ratio of 1:1, which further demonstrates that videos usually contain complex audios with both verbal and nonverbal information, while previous multimodal pre-training works have not specifically considered handling them simultaneously.
The results on the three datasets show that our audio type token can effectively control the information aspect of encoded audio features for different application scenarios.

\noindent\textbf{Key Components}
Table~\ref{tab:abalation_components_retrieval2} ablates the contributions from key components of our model.
The text-audio retrieval results on MSR-VTT and VATEX datasets consistently demonstrate the effectiveness of each proposed component.
Especially, compared with row 1, directly initializing the audio backbone with vision backbone~(row 2) has brought obvious gains, which further demonstrates that the audio information learning can benefit from existing visual knowledge.

\noindent\textbf{Training Cost}
Fully exploiting the existing vision-text knowledge for audio pre-training can not only help the audio representation learning, but also reduce the training cost. In this section we compare the training cost and the classification performance on ECS50 and  UCF101 with VATT~\cite{Akbari_VATT_arxiv21},  which is a vision-text-audio model pre-trained from scratch. 
For fair comparison, we follow the VATT to train a linear classifier on top of the frozen multimodal backbones, and report the mean accuracy over official splits~(5-fold and 3-fold cross validation for ESC50  and UCF101 respectively) .
As the results shown in Table~\ref{tab:training_cost}, our CLIP4VLA model achieves better downstream results with much less training cost, which demonstrates the advantages of learning audio from the existing visual-text knowledge.

\section{Conclusion}
We propose CLIP4VLA for Vision-Language-Audio processing by extending the VL pre-training model CLIP to accommodate the audio modality in a unified and economic way, which
incorporates an audio encoder with the same structure as the vision backbone for training consistency and efficiency.
To take full advantage of multimodal training data, we propose the contrastive learning from both inter- and intra-modal perspectives. 
Considering both verbal information and nonverbal information contained in general audios, we further propose an audio type token to explicitly encode these two types of information.
CLIP4VLA is validated by the video retrieval and video captioning tasks on MSR-VTT, VATEX, and Audiocaps benchmark datasets and achieves the state-of-the-art performance.

\section{Acknowledgments}
This work was partially supported by National Key R\&D Program of China (No. 2020AAA0108600) and National Natural Science Foundation of China (No. 62072462).
\balance
\bibliography{aaai23}

\begin{thebibliography}{46}
\providecommand{\natexlab}[1]{#1}

\bibitem[{Akbari et~al.(2021)Akbari, Yuan, Qian, Chuang, Chang, Cui, and
  Gong}]{Akbari_VATT_arxiv21}
Akbari, H.; Yuan, L.; Qian, R.; Chuang, W.; Chang, S.; Cui, Y.; and Gong, B.
  2021.
\newblock {VATT:} Transformers for Multimodal Self-Supervised Learning from Raw
  Video, Audio and Text.
\newblock In \emph{NeurIPS}.

\bibitem[{Alayrac et~al.(2020)Alayrac, Recasens, Schneider, Arandjelovic,
  Ramapuram, Fauw, Smaira, Dieleman, and Zisserman}]{Alayrac_MMV_NIPS20}
Alayrac, J.; Recasens, A.; Schneider, R.; Arandjelovic, R.; Ramapuram, J.;
  Fauw, J.~D.; Smaira, L.; Dieleman, S.; and Zisserman, A. 2020.
\newblock Self-Supervised MultiModal Versatile Networks.
\newblock In \emph{NeurIPS}.

\bibitem[{Baevski et~al.(2020)Baevski, Zhou, Mohamed, and
  Auli}]{Alexei_wav2vec2_NIPS20}
Baevski, A.; Zhou, Y.; Mohamed, A.; and Auli, M. 2020.
\newblock wav2vec 2.0: {A} Framework for Self-Supervised Learning of Speech
  Representations.
\newblock In \emph{NeurIPS}.

\bibitem[{Carion et~al.(2020)Carion, Massa, Synnaeve, Usunier, Kirillov, and
  Zagoruyko}]{Carion_DERT_ECCV20}
Carion, N.; Massa, F.; Synnaeve, G.; Usunier, N.; Kirillov, A.; and Zagoruyko,
  S. 2020.
\newblock End-to-End Object Detection with Transformers.
\newblock In \emph{ECCV}.

\bibitem[{Chen et~al.(2021)Chen, Hu, Wang, Zhou, and Li}]{Chen_CLIPFRL_ICCV21}
Chen, A.; Hu, F.; Wang, Z.; Zhou, F.; and Li, X. 2021.
\newblock What Matters for Ad-hoc Video Search? {A} Large-scale Evaluation on
  {TRECVID}.
\newblock In \emph{ICCV}.

\bibitem[{Chen et~al.(2022)Chen, Wang, Chen, Wu, Liu, Chen, Li, Kanda,
  Yoshioka, Xiao, Wu, Zhou, Ren, Qian, Qian, Wu, Zeng, and
  Wei}]{Chen_WavLM_arxiv21}
Chen, S.; Wang, C.; Chen, Z.; Wu, Y.; Liu, S.; Chen, Z.; Li, J.; Kanda, N.;
  Yoshioka, T.; Xiao, X.; Wu, J.; Zhou, L.; Ren, S.; Qian, Y.; Qian, Y.; Wu,
  J.; Zeng, M.; and Wei, F. 2022.
\newblock WavLM: Large-Scale Self-Supervised Pre-Training for Full Stack Speech
  Processing.
\newblock \emph{{IEEE} J. Sel. Top. Signal Process.}

\bibitem[{Chen et~al.(2020)Chen, Zhao, Jin, and Wu}]{Chen_HGR_CVPR20}
Chen, S.; Zhao, Y.; Jin, Q.; and Wu, Q. 2020.
\newblock Fine-Grained Video-Text Retrieval With Hierarchical Graph Reasoning.
\newblock In \emph{CVPR}.

\bibitem[{Chung et~al.(2019)Chung, Hsu, Tang, and
  Glass}]{Chung_APC_Interspeech19}
Chung, Y.; Hsu, W.; Tang, H.; and Glass, J.~R. 2019.
\newblock An Unsupervised Autoregressive Model for Speech Representation
  Learning.
\newblock In \emph{Interspeech}.

\bibitem[{Fang et~al.(2021)Fang, Xiong, Xu, and Chen}]{Fang_CLIP2Video_arxiv21}
Fang, H.; Xiong, P.; Xu, L.; and Chen, Y. 2021.
\newblock CLIP2Video: Mastering Video-Text Retrieval via Image {CLIP}.
\newblock \emph{CoRR}.

\bibitem[{Gabeur et~al.(2020)Gabeur, Sun, Alahari, and
  Schmid}]{Gabeur_MMT_ECCV20}
Gabeur, V.; Sun, C.; Alahari, K.; and Schmid, C. 2020.
\newblock Multi-modal Transformer for Video Retrieval.
\newblock In \emph{ECCV}.

\bibitem[{Gemmeke et~al.(2017)Gemmeke, Ellis, Freedman, Jansen, Lawrence,
  Moore, Plakal, and Ritter}]{Gemmeke_Audioset_ICASSP17}
Gemmeke, J.~F.; Ellis, D. P.~W.; Freedman, D.; Jansen, A.; Lawrence, W.; Moore,
  R.~C.; Plakal, M.; and Ritter, M. 2017.
\newblock Audio Set: An ontology and human-labeled dataset for audio events.
\newblock In \emph{ICASSP}.

\bibitem[{Gong, Chung, and Glass(2021)}]{Gong_AST_arxiv21}
Gong, Y.; Chung, Y.; and Glass, J.~R. 2021.
\newblock {AST:} Audio Spectrogram Transformer.
\newblock In \emph{Interspeech}.

\bibitem[{Guzhov et~al.(2020)Guzhov, Raue, Hees, and
  Dengel}]{Andrey_ESResnet_ICPR20}
Guzhov, A.; Raue, F.; Hees, J.; and Dengel, A. 2020.
\newblock ESResNet: Environmental Sound Classification Based on Visual Domain
  Models.
\newblock In \emph{ICPR}.

\bibitem[{Guzhov et~al.(2022)Guzhov, Raue, Hees, and
  Dengel}]{Guzhov_AudioCLIP_arxiv21}
Guzhov, A.; Raue, F.; Hees, J.; and Dengel, A. 2022.
\newblock Audioclip: Extending Clip to Image, Text and Audio.
\newblock In \emph{ICASSP}.

\bibitem[{Hsu et~al.(2021)Hsu, Bolte, Tsai, Lakhotia, Salakhutdinov, and
  Mohamed}]{Hsu_HuBERT_ACM21}
Hsu, W.; Bolte, B.; Tsai, Y.~H.; Lakhotia, K.; Salakhutdinov, R.; and Mohamed,
  A. 2021.
\newblock HuBERT: Self-Supervised Speech Representation Learning by Masked
  Prediction of Hidden Units.
\newblock \emph{{IEEE} {ACM} Trans. Audio Speech Lang. Process.}

\bibitem[{J{\'{o}}zefowicz et~al.(2016)J{\'{o}}zefowicz, Vinyals, Schuster,
  Shazeer, and Wu}]{Jozefowicz_NCEloss_arxiv16}
J{\'{o}}zefowicz, R.; Vinyals, O.; Schuster, M.; Shazeer, N.; and Wu, Y. 2016.
\newblock Exploring the Limits of Language Modeling.
\newblock \emph{CoRR}.

\bibitem[{Kim et~al.(2019)Kim, Kim, Lee, and Kim}]{Kim_audiocaps_NAACL19}
Kim, C.~D.; Kim, B.; Lee, H.; and Kim, G. 2019.
\newblock AudioCaps: Generating Captions for Audios in The Wild.
\newblock In \emph{NAACL-HLT}.

\bibitem[{Lei et~al.(2021)Lei, Li, Zhou, Gan, Berg, Bansal, and
  Liu}]{Lei_ClipBert_CVPR21}
Lei, J.; Li, L.; Zhou, L.; Gan, Z.; Berg, T.~L.; Bansal, M.; and Liu, J. 2021.
\newblock Less is More: ClipBERT for Video-and-Language Learning via Sparse
  Sampling.
\newblock In \emph{CVPR}.

\bibitem[{Li et~al.(2020)Li, Chen, Cheng, Gan, Yu, and Liu}]{Li_HERO_EMNLP20}
Li, L.; Chen, Y.; Cheng, Y.; Gan, Z.; Yu, L.; and Liu, J. 2020.
\newblock {HERO:} Hierarchical Encoder for Video+Language Omni-representation
  Pre-training.
\newblock In \emph{EMNLP}.

\bibitem[{Li et~al.(2019)Li, Xu, Yang, Chen, and Dong}]{Li_W2VV_MM19}
Li, X.; Xu, C.; Yang, G.; Chen, Z.; and Dong, J. 2019.
\newblock {W2VV++:} Fully Deep Learning for Ad-hoc Video Search.
\newblock In \emph{ACM MM}.

\bibitem[{Lin, Gan, and Wang(2021)}]{Lin_APML_AAAI21}
Lin, K.; Gan, Z.; and Wang, L. 2021.
\newblock Augmented Partial Mutual Learning with Frame Masking for Video
  Captioning.
\newblock In \emph{AAAI}.

\bibitem[{Ling et~al.(2020)Ling, Liu, Salazar, and
  Kirchhoff}]{Ling_Decoar_ICASSP20}
Ling, S.; Liu, Y.; Salazar, J.; and Kirchhoff, K. 2020.
\newblock Deep Contextualized Acoustic Representations for Semi-Supervised
  Speech Recognition.
\newblock In \emph{ICASSP}.

\bibitem[{Liu, Chung, and Glass(2021)}]{Liu_NPC_arxiv20}
Liu, A.~H.; Chung, Y.; and Glass, J.~R. 2021.
\newblock Non-Autoregressive Predictive Coding for Learning Speech
  Representations from Local Dependencies.
\newblock In \emph{Interspeech}.

\bibitem[{Liu, Li, and Lee(2021)}]{Liu_TERA_ACM21}
Liu, A.~T.; Li, S.; and Lee, H. 2021.
\newblock {TERA:} Self-Supervised Learning of Transformer Encoder
  Representation for Speech.
\newblock \emph{{IEEE} {ACM} Trans. Audio Speech Lang. Process.}

\bibitem[{Liu et~al.(2021)Liu, Zhu, Liu, Guo, Zhao, Sun, Wang, Lu, Zhou, Zhang,
  and Wang}]{Liu_OPT_arxiv21}
Liu, J.; Zhu, X.; Liu, F.; Guo, L.; Zhao, Z.; Sun, M.; Wang, W.; Lu, H.; Zhou,
  S.; Zhang, J.; and Wang, J. 2021.
\newblock {OPT:} Omni-Perception Pre-Trainer for Cross-Modal Understanding and
  Generation.
\newblock \emph{CoRR}.

\bibitem[{Liu et~al.(2019)Liu, Albanie, Nagrani, and Zisserman}]{Liu_CE_BMVC19}
Liu, Y.; Albanie, S.; Nagrani, A.; and Zisserman, A. 2019.
\newblock Use What You Have: Video retrieval using representations from
  collaborative experts.
\newblock In \emph{BMVC}.

\bibitem[{Luo et~al.(2020)Luo, Ji, Shi, Huang, Duan, Li, Li, Bharti, and
  Zhou}]{luo_univl_arxiv20}
Luo, H.; Ji, L.; Shi, B.; Huang, H.; Duan, N.; Li, T.; Li, J.; Bharti, T.; and
  Zhou, M. 2020.
\newblock UniVL: A Unified Video and Language Pre-Training Model for Multimodal
  Understanding and Generation.
\newblock arXiv:2002.06353.

\bibitem[{Luo et~al.(2021)Luo, Ji, Zhong, Chen, Lei, Duan, and
  Li}]{Luo_CLIP4Clip_arxiv21}
Luo, H.; Ji, L.; Zhong, M.; Chen, Y.; Lei, W.; Duan, N.; and Li, T. 2021.
\newblock CLIP4Clip: An Empirical Study of {CLIP} for End to End Video Clip
  Retrieval.
\newblock \emph{CoRR}.

\bibitem[{Miech, Laptev, and Sivic(2018)}]{Miech_moee_arxiv18}
Miech, A.; Laptev, I.; and Sivic, J. 2018.
\newblock Learning a Text-Video Embedding from Incomplete and Heterogeneous
  Data.
\newblock \emph{CoRR}.

\bibitem[{Miech et~al.(2019)Miech, Zhukov, Alayrac, Tapaswi, Laptev, and
  Sivic}]{Miech_Howto100m_ICCV19}
Miech, A.; Zhukov, D.; Alayrac, J.; Tapaswi, M.; Laptev, I.; and Sivic, J.
  2019.
\newblock HowTo100M: Learning a Text-Video Embedding by Watching Hundred
  Million Narrated Video Clips.
\newblock In \emph{ICCV}.

\bibitem[{Patrick et~al.(2021)Patrick, Huang, Asano, Metze, Hauptmann,
  Henriques, and Vedaldi}]{Patrick_SSB_ICLR_21}
Patrick, M.; Huang, P.; Asano, Y.~M.; Metze, F.; Hauptmann, A.~G.; Henriques,
  J.~F.; and Vedaldi, A. 2021.
\newblock Support-set bottlenecks for video-text representation learning.
\newblock In \emph{ICLR}.

\bibitem[{Piczak(2015)}]{Piczak_ESC50_mm15}
Piczak, K.~J. 2015.
\newblock {ESC:} Dataset for Environmental Sound Classification.
\newblock In \emph{ACM MM}.

\bibitem[{Portillo{-}Quintero, Ortiz{-}Bayliss, and
  Terashima{-}Mar{\'{\i}}n(2021)}]{Jes_CLIP_MCPR2021}
Portillo{-}Quintero, J.~A.; Ortiz{-}Bayliss, J.~C.; and
  Terashima{-}Mar{\'{\i}}n, H. 2021.
\newblock A Straightforward Framework for Video Retrieval Using {CLIP}.
\newblock In \emph{MCPR}.

\bibitem[{Soomro, Zamir, and Shah(2012)}]{Khurram_UCF101_2012}
Soomro, K.; Zamir, A.~R.; and Shah, M. 2012.
\newblock {UCF101:} {A} Dataset of 101 Human Actions Classes From Videos in The
  Wild.
\newblock \emph{CoRR}.

\bibitem[{Sun et~al.(2020)Sun, Baradel, Murphy, and Schmid}]{Sun_CBT_ECCV20}
Sun, C.; Baradel, F.; Murphy, K.; and Schmid, C. 2020.
\newblock Learning Video Representations using Contrastive Bidirectional
  Transformer.
\newblock In \emph{ECCV}.

\bibitem[{Sun et~al.(2019)Sun, Myers, Vondrick, Murphy, and
  Schmid}]{Sun_videoBERT_ICCV19}
Sun, C.; Myers, A.; Vondrick, C.; Murphy, K.; and Schmid, C. 2019.
\newblock VideoBERT: {A} Joint Model for Video and Language Representation
  Learning.
\newblock In \emph{ICCV}.

\bibitem[{Tang, Lei, and Bansal(2021)}]{Tang_DeCEMBERT_NAACL21}
Tang, Z.; Lei, J.; and Bansal, M. 2021.
\newblock DeCEMBERT: Learning from Noisy Instructional Videos via Dense
  Captions and Entropy Minimization.
\newblock In \emph{NAACL-HLT}.

\bibitem[{van~den Oord, Li, and Vinyals(2018)}]{Oord_CPC_arxiv18}
van~den Oord, A.; Li, Y.; and Vinyals, O. 2018.
\newblock Representation Learning with Contrastive Predictive Coding.
\newblock \emph{CoRR}.

\bibitem[{Wang et~al.(2022)Wang, Lin, Hoi, and Miao}]{Wang_CMG_arxiv21}
Wang, H.; Lin, G.; Hoi, S. C.~H.; and Miao, C. 2022.
\newblock Cross-Modal Graph With Meta Concepts for Video Captioning.
\newblock \emph{{IEEE} Trans. Image Process.}

\bibitem[{Wang et~al.(2019)Wang, Wu, Chen, Li, Wang, and
  Wang}]{Wang_Vatex_ICCV19}
Wang, X.; Wu, J.; Chen, J.; Li, L.; Wang, Y.; and Wang, W.~Y. 2019.
\newblock VaTeX: {A} Large-Scale, High-Quality Multilingual Dataset for
  Video-and-Language Research.
\newblock In \emph{ICCV}.

\bibitem[{Wu et~al.(2022)Wu, Seetharaman, Kumar, and
  Bello}]{Wu_Wav2CLIP_arxiv21}
Wu, H.; Seetharaman, P.; Kumar, K.; and Bello, J.~P. 2022.
\newblock Wav2CLIP: Learning Robust Audio Representations from Clip.
\newblock In \emph{ICASSP}.

\bibitem[{Xie et~al.(2018)Xie, Sun, Huang, Tu, and Murphy}]{Xie_S3D_ECCV18}
Xie, S.; Sun, C.; Huang, J.; Tu, Z.; and Murphy, K. 2018.
\newblock Rethinking Spatiotemporal Feature Learning For Video Understanding.
\newblock In \emph{ECCV}.

\bibitem[{Xu et~al.(2021)Xu, Ghosh, Huang, Arora, Aminzadeh, Feichtenhofer,
  Metze, and Zettlemoyer}]{Xu_VLM_ACL21}
Xu, H.; Ghosh, G.; Huang, P.; Arora, P.; Aminzadeh, M.; Feichtenhofer, C.;
  Metze, F.; and Zettlemoyer, L. 2021.
\newblock {VLM:} Task-agnostic Video-Language Model Pre-training for Video
  Understanding.
\newblock In \emph{ACL}.

\bibitem[{Xu et~al.(2016)Xu, Mei, Yao, and Rui}]{Xu_MSRVTT_CVPR16}
Xu, J.; Mei, T.; Yao, T.; and Rui, Y. 2016.
\newblock {MSR-VTT:} {A} Large Video Description Dataset for Bridging Video and
  Language.
\newblock In \emph{CVPR}.

\bibitem[{Zhang et~al.(2020)Zhang, Shi, Yuan, Li, Wang, Hu, and
  Zha}]{Zhang_ORG_CVPR20}
Zhang, Z.; Shi, Y.; Yuan, C.; Li, B.; Wang, P.; Hu, W.; and Zha, Z. 2020.
\newblock Object Relational Graph With Teacher-Recommended Learning for Video
  Captioning.
\newblock In \emph{CVPR}.

\bibitem[{Zhu and Yang(2020)}]{Zhu_ActBERT_CVPR20}
Zhu, L.; and Yang, Y. 2020.
\newblock ActBERT: Learning Global-Local Video-Text Representations.
\newblock In \emph{CVPR}.

\end{thebibliography}
\clearpage
\appendix
\section{Appendix}
This document provides the supplementary materials that have been omitted from the main paper due to space limitations. Section~\ref{sec:dataset} presents the detailed introduction of all related datasets and how we pre-process them.
Section~\ref{sec:implement_details} describes the experimental setup for both pre-training and fine-tuning.  
Section~\ref{sec:modality_fusion} analyzes the  effects of multimodal fusion methods and the contribution of each modality during fine-tuning.
In Section~\ref{sec:case_study}, we visualize the caption results of CLIP4VLA.

\subsection{More Dataset Details}
\label{sec:dataset}
This section further introduces the characteristics, scale, and the splits of our pre-training datasets~(Howto100M and Audioset), fine-tuning datasets~(MSR-VTT, VATEX, Audiocaps), and how we use them during experiments.\\

\noindent\textbf{Howto100M and Audioset}
We use Howto100M and Audioset as our pre-training datasets. 
Since the originally released Howto100M only consists of silent videos and transcripts from ASR, we crawl their corresponding audios from YouTube and filter out the mismatch and unavailable ones. 
Finally, we get 0.95M videos with an average duration of 6.5 minutes and an average clip-text pair of 110 per video. 
Audioset is a collection of over 2 million 10-second video clips, each clip is labeled with an event label from a set of 527 labels.
To generate coherent sentences from discrete labels,  
we use the template of ``The sound of \underline{~~~},\underline{~~~},$\cdots$'' and fill in the blanks with event annotations. 
After filtering out the validation data and unavailable data in Audioset, we finally get 1.6 million video-text-audio pairs for training. 
In practice, we also deal with the mismatch of duration and quantity between these two datasets. For the duration mismatch, we extend the video clip of Howto100M from an average of 6s to at least 10s to keep consistent with Audioset via concatenating the clip with its neighbor clips.
For the quantity mismatch,  we keep the 1:1 data ratio of these two datasets within a batch.

\vspace{6pt}
\noindent\textbf{MSR-VTT}
MSR-VTT is a visual-centric open domain dataset for text-video retrieval and video captioning. Each video has 20 manually labeled caption sentences. For retrieval, we use ``The training-9k''\& ``The testing-1k'' as the default setting, and for captioning, we use the MSR-VTT’s standard split (MSR-VTT
Training-6K), i.e.  6512, 498, and 2990 clips for training, validation, and testing.
After filtering out videos without audios, video retrieval remains 7867 and 884 videos for training and testing respectively. Video captioning remains 5867, 448, and 2617 videos for training, validation, and testing respectively.

\vspace{6pt}
\noindent\textbf{VATEX}
VATEX is a large-scale visual-centric dataset that reuses a subset of the videos from the Kinetics-600.  Each video is annotated with 10 English and 10 Chinese descriptions, and only the English corpora are used in our experiments. 
Following the official data splits and after filtering out silent videos, There are in total 24667, 1424, 1421 videos for training, validation, and testing respectively for the video retrieval task,  and 24667, 2845, 5698 videos for training, validation, testing respectively for the video captioning task.

\vspace{6pt}
\noindent\textbf{Audiocaps}
Audiocaps is an audio-centric video dataset created based on Audioset, whose audios are mainly in event scenarios with durations shorter than 10 seconds.
It is divided into three splits, and each video clip in the training set contains one caption, while five captions per clip are used in validation and testing sets. After filtering out videos that YouTube-hosted source is no longer available, we finally use 49712, 495, 967 videos for training, validation, and testing respectively for the retrieval task.

\subsection{ Implementation Details}
\label{sec:implement_details}
In this section, we provide more details of the adjustments above VL pre-trained model CLIP, tri-modal input pre-processing and the hyper-parameters settings during experiments.

To make full use of existing visual-language knowledge, we initialize the text backbone and vision backbone with CLIP, ViT-B/32 version, and  initialize the audio backbone with the vision backbone. For the cross encoder used during fine-tuning, we initialize it with the first 4 layers of the text encoder, except for the type embeddings $P$. 
In addition, we make the following  adjustments to adapt CLIP to CLIP4VLA:
(1) As the cross encoder is initialized from the text encoder, which contains a single-direction attention mask to block the interaction among each token with its subsequent ones, we delete it during the retrieval fine-tuning.
(2) Expand the positional embedding of the audio backbone. Since we add an extra audio type token at the end of the audio patch sequence, we randomly initialize the corresponding positional embedding not included in CLIP during pre-training, and repeat it twice for fine-tuning.
Other extra  parameters, including the audio type tokens, linear layer for retrieval fine-tuning~(similarity calculation in Global to Local, Local to Local) and caption fine-tuning, are randomly initialized.

\begin{table*}[t]
\centering
\caption{Comparison of different modal fusion methods for video retrieval on Audiocaps, MSR-VTT and VATEX. G2G, G2L, L2L are short for multimodal fusion methods of \emph{Global to Global}, \emph{Global to Local}, and \emph{Local to Local} respectively}
  \begin{tabular}{ccccccccccccccccc}
    \toprule
   ~& ~ &\multicolumn{4}{c}{Audiocaps} && \multicolumn{4}{c}{MSR-VTT} && \multicolumn{4}{c}{VATEX}  \\
    \cline{3-6} \cline{8-11} \cline{13-16} 
    ~& Method & R@1 & R@5 & R@10 & MR & & R@1 & R@5 & R@10 & MR& & R@1 & R@5 & R@10 & MR \\
     \midrule
     \multirow{3}{*}{\rotatebox[origin=c]{90}{V}}& G2G &12.3&37.1&52.7&9.0&&\textbf{44.1} &\textbf{71.9} & \textbf{81.9} & \textbf{ 2.0}&&57.5&87.9&93.9& 1.0\\
     ~& G2L&\textbf{12.4}&\textbf{37.4}&\textbf{53.1}&\textbf{9.0}&&43.2&69.3&81.0&2.0&&\textbf{60.1}& \textbf{89.7}& \textbf{94.9}& \textbf{1.0}\\
     ~& L2L& 11.2 & 34.2 & 49.8 & 11.0&&39.5&71.0&80.9&2.0&&52.6&85.9&92.8&1.0\\
      \midrule
      \multirow{3}{*}{\rotatebox[origin=c]{90}{A}}& G2G&26.8&59.7&73.9&4.0&&\textbf{10.6}&\textbf{26.5}&\textbf{38.0}&\textbf{19.0}&&7.4&21.4& 20.2& 35.0\\
      ~&G2L &27.8&60.9&75.8&4.0&&10.2&24.7&35.8&   21.0&&\textbf{9.6}&\textbf{24.8}&\textbf{34.2}&\textbf{27.0}\\
      ~&L2L&\textbf{28.4}&\textbf{60.9}&\textbf{76.2}&  \textbf{4.0}&&10.1&24.6&33.7&24.0&&7.9&22.1&30.8&35.0\\
      \midrule
      \multirow{3}{*}{\rotatebox[origin=c]{90}{V\&A}}&G2G &31.7&66.3&81.4&  3.0 && 44.5&73.3&82.9& 2.0 && 60.5 & 89.9 & 95.1 &  1.0\\
      ~&G2L&33.4&\textbf{68.4}&81.9& 3.0&&\textbf{46.8}&\textbf{73.7}&\textbf{84.0}& \textbf{2.0}&& \textbf{62.8}&\textbf{91.1}&\textbf{95.9}&   \textbf{1.0}\\
      ~&L2L & \textbf{33.6} & 68.1 & \textbf{82.3} & \textbf{3.0} &&42.8& 71.8&80.1& 2.0&&54.8&87.3& 94.0& 1.0\\
    \bottomrule
\end{tabular}
\label{tab:abalation_modal_fuse_retrieval}
\end{table*}

\begin{table*}[t]
\caption{Comparison of video captioning performance with different modalities on MSR-VTT and VATEX  }

\centering
  \begin{tabular}{ccccccccccc}
    \toprule
    ~ &\multicolumn{4}{c}{MSR-VTT} && \multicolumn{4}{c}{VATEX}  \\
    \cline{2-5} \cline{7-10}  
     Modality & BLUE4 & METEOR & ROUGE & CIDER & & BLUE4 & METEOR & ROUGE & CIDER \\
     \midrule
      V&  44.4& 30.1& 62.9&55.7&&35.0 & 24.6 &53.9 & 58.5  \\
      A & 32.1&23.9&54.6&27.5&&17.7&16.4&42.7&16.1\\
      V\&A &47.8&31.1&64.4&58.0&&36.2&24.9&54.5&59.8\\
    \bottomrule
\end{tabular}
\label{tab:abalation_modal_fuse_caption}
\end{table*}

For the visual input, we uniformly sample at most 12 frames from each video/video clip at 1 fps.
For the audio input, we firstly downsample the audio wave to 16000 Hz, then transform it into a spectrogram, and further split it into segment sequence. 
According to the maximum video length of each dataset, we set fixed number of audio segments $L^A=16$~(around 27 seconds) for MSR-VTT, and $L^A=6$~(around 11 seconds) for other datasets~(Howto100m, Audioset, Audiocaps, VATEX).
To get the audio segments with continuous information, we sample the audio segments from the middle of the video to both sides.
The sentences are tokenized and padded to length 32 for retrieval and 48 for captioning.
During pre-training, we use Adam as the optimizer, 1e-6 as the learning rate, and
cosine schedule as the learning rate decay strategy following the setup of CLIP.
The overall training procedure takes 48 V100 days, 300k steps with batchsize of 256.
During fine-tuning, we set the learning rate as le-7 for all pre-trained parameters~(e.g. 3 backbones), and as 5e-4 for other additional parameters~(e.g. cross encoder).
We fix the vision backbone and text backbone during pre-training and release all other parameters during fine-tuning.
For video retrieval, we set batchsize as 128 for modality fusion methods of \textit{Global to Global} and \textit{Global to Local}, 32 for \textit{Local to Local} since it requires more memory to interact text with each video.

\subsection{Modality Fusion}
\label{sec:modality_fusion}
In this section, we analyze the contributions of each modality in video retrieval (Table~\ref{tab:abalation_modal_fuse_retrieval}) and video captioning tasks (Table~\ref{tab:abalation_modal_fuse_caption}), and compare different multimodal fusion methods (\emph{Global to Global}, \emph{Global to Local} and \emph{Local to Local}) for video retrieval.

Although the importance degree of vision and audio varies under different scenarios, combining the two modalities always achieves better performances than the single modality for both video retrieval and captioning tasks.
To further analyze the multimodal interactions, we compare different multimodal fusion methods in Table~\ref{tab:abalation_modal_fuse_retrieval}, where the results vary on different datasets.
Specifically, the \emph{Local to Local} approach works the best for Audiocaps dataset, because the temporal information of the text query is crucial to describe the event order in videos~(e.g. ``A woman singing then choking followed by birds chirping''), while the \emph{Local to Local} method can capture this temporal semantic by word-level interactions.
For the MSR-VTT and VATEX datasets, whose text descriptions are more general, the temporal information of vision and audio modalities matters more, which can be extracted by \emph{Global to Local}. 
An unexpected phenomenon is that the \emph{Global to Local} performs worse than \emph{Global to Global} in single modal retrieval on MSR-VTT. We think it is because the extra cross transformer requires sufficient training data, while 7,867 training videos in MSR-VTT is not enough. Adding an extra modality~($8^{th}$ row on MSR-VTT) or enlarging the training set~(results of VATEX) could mitigate this problem.

\subsection{Qualitative Results}
\label{sec:case_study}
\begin{figure*}[t]
  \centering
    \caption{Caption results of CLIP4VLA with different modal inputs. \textbf{V} refers to vision modality and \textbf{A} refers to audio modality. The red represents the nonverbal information in the audio, and the blue represents the verbal information }
  \includegraphics[width=\textwidth]{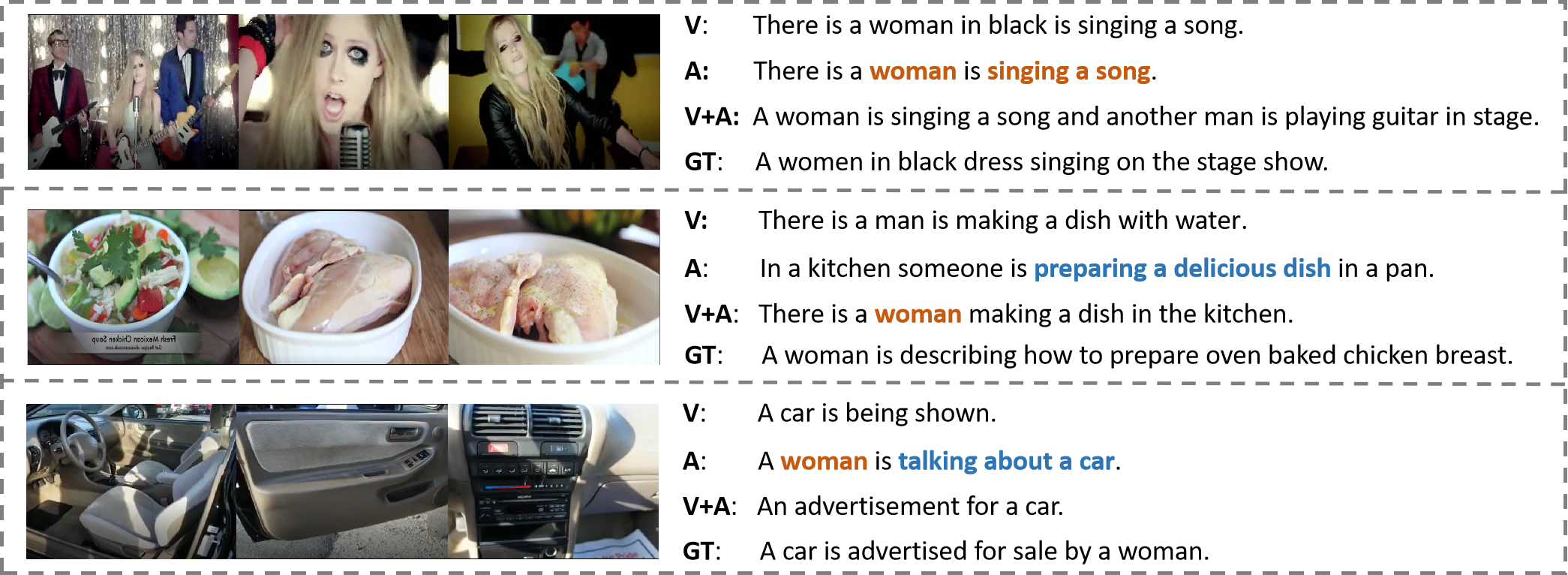}

  \label{fig:case_study}
\end{figure*}

In Figure~\ref{fig:case_study}, we visualize the caption results of our proposed CLIP4VLA with different modal inputs on MSR-VTT. 
In the first example, our model successfully extracts the nonverbal information of the audio so that it generates ``woman'', ``singing a song'' in the caption. 
The second example refers to a  cooking tutorial video with no characters in the image. CLIPVLA recognizes the cooking scene from the spoken language, which demonstrates its ability on verbal information extraction. 
The third video is a car advertisement with a car's image and a woman's introduction. With the help of audio information,  our model correctly understands the video as an ``advertisement'', rather than ``A car is being shown''.

The above cases further demonstrate that our proposed CLIP4VLA can extract both verbal and nonverbal information in the audio. However, CLIP4VLA can't recognize the finer-grained verbal information in speech, so it fails to describe the cooking details of the second tutorial and the car's details of the third advertisement. We will continue to improve its ability on fine-grained verbal information extraction in our future work.

\end{document}